\documentclass[letterpaper, 10 pt, conference]{ieeeconf}  

\IEEEoverridecommandlockouts                              

\usepackage{graphicx} 
\usepackage{amsmath} 
\usepackage{amssymb}  
\usepackage{multirow}
\usepackage{makecell}
\usepackage[colorinlistoftodos]{todonotes}

\usepackage{xcolor}
\usepackage{url}

\usepackage{tikz}
\usepackage{textcomp}
\usepackage{hyperref}
\usepackage{lipsum}

\newcommand\copyrighttext{%
  \footnotesize \textcopyright 2021 IEEE. Personal use of this material is permitted.
  Permission from IEEE must be obtained for all other uses, in any current or future
  media, including reprinting/republishing this material for advertising or promotional
  purposes, creating new collective works, for resale or redistribution to servers or
  lists, or reuse of any copyrighted component of this work in other works.}
\newcommand\copyrightnotice{%
\begin{tikzpicture}[remember picture,overlay]
\node[anchor=south,yshift=10pt] at (current page.south) {\fbox{\parbox{\dimexpr\textwidth-\fboxsep-\fboxrule\relax}{\copyrighttext}}};
\end{tikzpicture}%
}

\title{\LARGE \bf
 On the descriptive power of LiDAR intensity images \\ for segment-based loop closing in 3-D SLAM
}

\author{Jan Wietrzykowski$^{1}$ and Piotr Skrzypczy\'{n}ski$^{1}$
\thanks{*This work was supported by the National Science Centre (NCN), contract no. UMO-2018/31/N/ST6/00941.
The work of P. Skrzypczy{\'n}ski was supported by TAILOR, a project funded by EU Horizon 2020 research and innovation program under GA No. 952215}
\thanks{$^{1}$Jan Wietrzykowski and Piotr Skrzypczy\'{n}ski are with
          Institute of Robotics and Machine Intelligence,
          Poznan University of Technology,
          Piotrowo 3A, 60-965 Pozna\'{n}, Poland
        {\tt\small name.surname@put.poznan.pl}}%
}

\DeclareMathOperator*{\argmin}{arg\,min}

\begin{document}

\copyrightnotice

\maketitle
\thispagestyle{empty}
\pagestyle{empty}

\begin{abstract}
We propose an extension to the segment-based global localization method for LiDAR SLAM
using descriptors learned considering the visual context of the segments.
A new architecture of the deep neural network is presented that learns the visual context acquired from synthetic LiDAR intensity images. This approach allows a single multi-beam LiDAR to produce rich and highly descriptive location signatures. The method is tested on two public datasets, demonstrating an improved descriptiveness of the new descriptors, and more reliable loop closure detection in SLAM. Attention analysis of the network is used to show the importance of focusing on the broader context rather than only on the 3-D segment.
\end{abstract}

\section{INTRODUCTION}

3-D laser SLAM methods recently became one of the key components of autonomous vehicles.
A majority of them are based on variants of the ICP (Iterative Closest Points) concept, i.e. on matching laser points to other points or ad-hoc created local structures, such as planes and line segments~\cite{sensors2021,loam,loamvariant}.
The accuracy of such methods is sufficient for most navigational tasks, thus the development is focused on making them more reliable and robust by giving an ability to recover from failures and to correct drift when the same location is revisited.
Unfortunately, working on the level of points makes the tasks of loop closure and re-localization difficult due to a lack of discriminative features that could be matched between temporarily distant observations.
On the other hand, point cloud retrieval, like in \cite{pointnetvlad}, provides only place recognition, not metric localization.
A different approach to laser SLAM is to cluster the point clouds into larger segments representing meaningful objects or their parts, e.g. cars, trees, parts of buildings~\cite{Dube2020}.
Those segments can be then matched between the current observation and the map, enabling metric global localization used for loop closing and re-localization.
However, relying solely on the geometry of isolated objects leaves a lot of uncertainty, because there could be many objects with similar shapes e.g. trees, walls.
We conjecture that it would be beneficial for the descriptiveness of those segments to include also information about the texture and the surroundings, which together constitute a broader (visual) context.

Using a single sensor is practical in applications, as it does not require accurate calibration that is necessary if a LiDAR-camera pair is employed \cite{nowicki2020}. 
However, it is problematic to obtain a visual context in LiDAR-only perception.
Fortunately, modern LiDARs provide also information about the intensity of the reflected beam.
Recent results \cite{zywanowski2020} suggest, that LiDAR intensity is more reliable for place recognition than regular camera images in some scenarios.
The nearer the wavelength to the visible light spectrum, the closer the readout to a passive camera image~\cite{ouster}.
This comes in handy when multi-beam LiDARs are used, which provide a relatively dense scan of the scene.
By arranging intensity readouts in a regular grid on a cylinder around the LiDAR, this information can be treated similarly to a camera image.
The intensity image carries a lot of information, not only about the appearance of the segment it may depict but also about the segment's surroundings.
Thus, it can be used to retrieve the visual context.

\begin{figure}[!t!]
    \centering
    \includegraphics[width=\columnwidth]{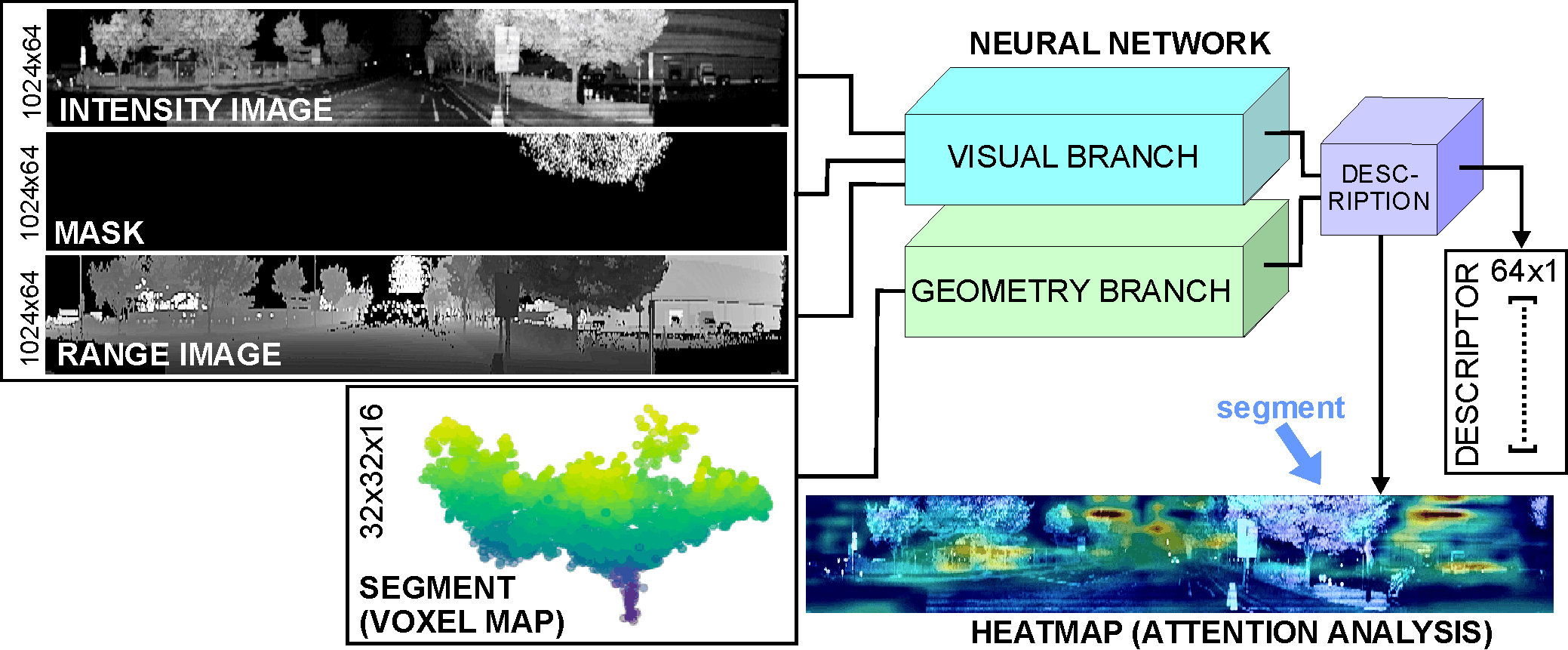}
    \caption{Concept of enhancing descriptors of 3-D segments by exploiting the broader visual context acquired from LiDAR intensity images.
    The included attention map shows that the network focuses on the context (warmer colors) rather than on the masked segment (pointed by the blue arrow).}
    \label{fig:idea}
\end{figure}

This paper attempts to bridge the gap that exists between the 3-D segment-based approach to loop closing, which turned out to be very practical in LiDAR SLAM \cite{loamvariant}, and the appearance-based approach, which is commonly applied in visual SLAM.
To our knowledge, we are the first to use intensity images to enhance the learned descriptors of 3-D segments.
We are also not aware of any work that researched learning description of segments visible in images.
Most papers tackle the problem of describing points and their local surrounding or the problem of global image descriptor computation.
Our solution falls in-between these two extremes, learning to describe segments that occupy part of the image, while also including the context in the description (Fig.~\ref{fig:idea}).
The contribution of this paper is as follows \footnote{Code available here: \url{https://github.com/LRMPUT/segmap_vis_views}}:
\begin{itemize}
    \item A method using intensity images to enhance segment descriptors (Section \ref{sec:visinp}).
    \item A novel architecture and training methodology for learning descriptors (Sections \ref{sec:arch} and \ref{sec:learn}).
    \item A procedure for attention analysis of the neural network in a case where output is a descriptor for in-depth analysis of our method (Section \ref{sec:attention}).
    \item Experimental verification of these methods (Section \ref{sec:perf}).    
\end{itemize}

\section{RELATED WORK}
In the field of LiDAR-based global localization, the use of segments is not broadly explored.
SegMatch, the predecessor of SegMap, introduced incremental segment growing and used hand-crafted features based on eigenvalues during matching~\cite{Dube2017}.
Tinchev \emph{et al.}~\cite{Tinchev2019} modified SegMap to use different descriptors learned by a lightweight network with X-Conv operations.

A significantly different approach to global localization using LiDAR measurements was presented by Chen \emph{et al.}~\cite{Chen2020}.
They used images of cylindrical projections in a similar way to our solution, but estimated overlap and yaw angle between a pair of views.
Contrary to our segment descriptor, they use a global descriptor of the whole image as an input to regression branches.
As this method outputs only similarity measure and relative yaw, ICP registration is needed to compute the 2-D pose.
Handcrafted global descriptors of LiDAR scans were used in LocNet~\cite{Yin2020}, but machine learning was applied to compare those descriptors.

Most of the work on describing appearance comes from the computer vision community and focuses on camera images.
SuperPoint~\cite{Detone2018} is one of the recent examples of learned detectors and descriptors.
The description is learned using warped images by minimizing hinge loss between all pairs of pixels.
On the other hand, in~\cite{Loquercio2017} descriptor projections are learned using triplet loss with L2 distance and hard negative mining.
The authors of~\cite{Loquercio2017} also show that including the context in a form of descriptors of larger portions of the image around the keypoint is beneficial for matching.
However, they provide only limited experiments with whole images to describe the context.
When it comes to range data, PointNetVLAD \cite{pointnetvlad} was the first DNN-based method producing a discriminative global descriptor for the localization task cast as large scale 3-D point cloud retrieval.
Using a similar approach to global localization with 3-D point clouds, PCAN \cite{zhang2019} introduced an attention mechanism that predicts significance of each point using its local geometrical context, but not the visual information.

The concept of using intensity information in global localization was explored only in a few studies.
Synthesized intensity images compared favorably in \cite{zywanowski2020} to regular camera images for DNN-based place recognition under varying weather conditions.
If handcrafted features are employed, as in~\cite{Cop2018}, where histograms of intensity were used,
the geometric information is not embedded in the descriptors and used only for hypothesis verification.
Histograms were also used by Guo \emph{et al.} \cite{Guo2019} in descriptors called ISHOT, which combined the SHOT descriptors and histograms of intensity differences around the keypoints.
When only local displacement is estimated, it is possible to exhaustively search the space of possible solutions.
L3Net~\cite{Lu2019} exploits cost volume spanning ($x$,$y$,yaw) space and directly minimizes the displacement error during training.
However, the description is done by detecting keypoints and describing local patches using coordinates and intensity values by a simple multi-layer perceptron.

\section{LOCALIZATION FRAMEWORK}
\label{sec:loc}

\begin{figure}[!t!]
    \centering
    \includegraphics[width=\columnwidth]{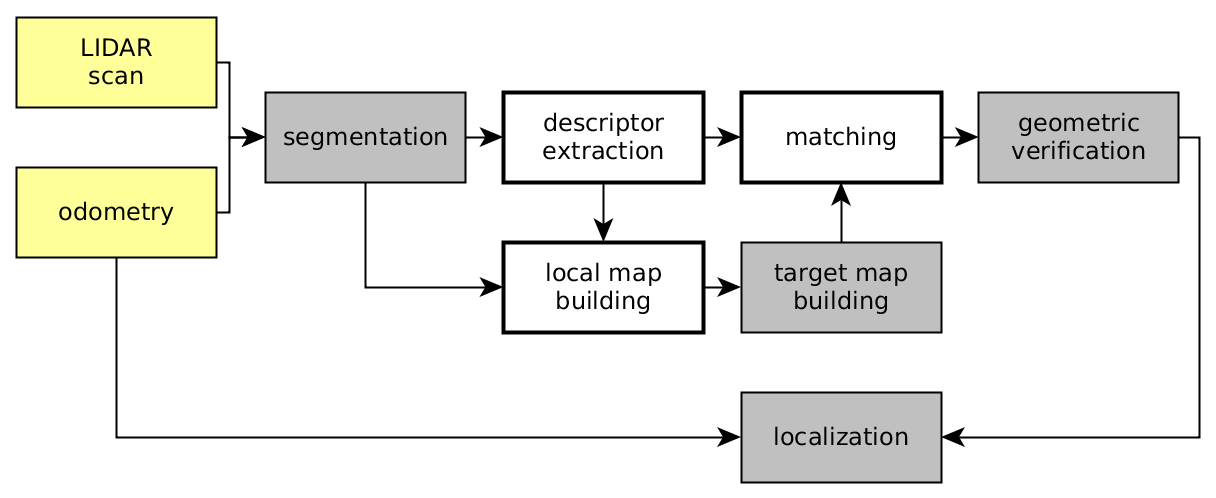}
    \caption{An overview of the localization pipeline in SegMap with marked modified modules (bolded, white rectangles) and unmodified modules (gray rectangles).}
    \label{fig:segmap}
\end{figure}

We demonstrate our novel approach to the context-aware description of 3-D segments extending the
open-source, modular SegMap framework \cite{Dube2020} for mapping and global localization.
Our solution inherits from SegMap the processing pipeline of the segment's geometry (clustering into segments)
and the general structure of the global map.
In the framework, we plug-in the new procedure of learning the descriptors, together with a
modified deep neural network (DNN) architecture, and a new segment matching procedure,
which better exploits the enhanced descriptors (Fig.~\ref{fig:segmap}).

The processing starts with an acquisition of a new LiDAR scan.
The scan is then matched to the previous one to estimate a sensor's displacement.
The displacement is integrated with the previous pose estimate and produces odometry,
but is also used to compensate the sensor motion by transforming all points to the common frame of reference \cite{loam}.
The local map of segments is obtained using the SegMap clustering procedures (incremental Euclidean segmenter) and the odometry estimates.
In comparison to the original SegMap, we also store synthesized intensity and range images in the local map.
Next, segments in the local map are described, which is the main focus of this paper.
The sensor pose estimate with respect to the target map is computed by finding match candidates for each segment's descriptor in the local map.
The candidates are being found using kNN search in the descriptor space among the descriptors of segments from the target map.
From all match candidates, the best subset is selected through consistency clustering and the final pose is computed using this subset.
As in SegMap, the target map can be either loaded from the disk or built along the trajectory by accumulating local map segments with their descriptors using current pose estimates.

\section{DESCRIPTION OF SEGMENTS}
\label{sec:desc}
Segments in the proposed solution are described using only LiDAR data, however, they owe their descriptiveness to the use of range readouts along with intensity data.
A local voxel grid representation of the segment is built from the range data, like in SegMap,
whereas the intensity data is converted to an image that provides a camera-like view of a segment.
An intensity image can be synthesized by directly exploiting the arrangement of measurement directions, as in the case of Ouster LiDAR,
or by a projection of the acquired point cloud onto a cylinder surrounding the sensor if the arrangement of the used readouts is not grid-like (e.g. from a Velodyne sensor).

\subsection{Visual input from a LiDAR}
\label{sec:visinp}
\begin{figure}[!t!]
    \centering
    \includegraphics[width=\columnwidth]{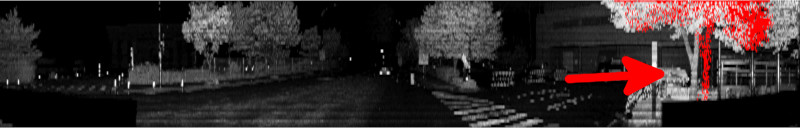}
    \caption{Exemplary mask misalignment caused by ignoring motion compensation of LiDAR scans.}
    \label{fig:mask}
\end{figure}

To describe the segment, an intensity image with the largest area of the projected segment is selected from available visual views and fed to the DNN along with the voxel grid representation of the segment.
Special care has to be taken to ensure that binary masks, denoting positions of segments in images, are aligned with objects in those images.
It is not feasible to track the origin of every point in a point cloud representing a segment, because of multiple filtering operations and the associated high computational and memory requirements.
Thus, we decided to compute masks by projecting point clouds onto images.
The simplest approach would be to project points onto a cylinder around the scanner as follows:
\begin{equation}
    r = \argmin_{r'} |\alpha_{r'} - \alpha |, \mbox{~~~}
    c = \mathrm{round} \left( \frac{\beta}{2 \pi} \cdot 1024 \right),
\end{equation}
where $(r, c)$ are resultant pixel coordinates, $r'$ iterate over LiDAR's scanning rings, $\alpha$ is an inclination angle of the point, $\beta$ is an azimuth angle of the point, and $\alpha_{r'}$ is an inclination angle for the ring $r'$.
However, a mask computed this way would be inaccurate, because measurements of pixels were not taken at the same time like in a global shutter camera.
Moreover, the time it takes a LiDAR to do a full scan is considerably longer than in a regular rolling shutter camera.
When this fact is ignored and points are projected onto the image surface using only one pose of the LiDAR, masks can be misaligned by a large margin, depending on a velocity (Fig.~\ref{fig:mask}).
We deal with this problem by keeping directions of rays coming out from the LiDAR, transformed to a common frame of reference by using displacement estimated by the odometry procedure.
Then, during projection, for every point $\mathbf{p}$, we choose the pixel $(i, j)$ with the closest direction $\mathbf{n}_{ij}$ of the scanning ray.
To speed up computations, we search only in a vicinity of the pixel $(r, c)$ that would be selected by the simple projection:
\begin{equation}
    (i^\ast, j^\ast) = \argmin_{\substack{r-r_m \le i < r+r_m\\ c-c_m \le j < c+c_m}} 1 - \arccos \frac{\mathbf{n}_{ij} \cdot \mathbf{p}}{|\mathbf{p}|},
\end{equation}
where $(i^\ast, j^\ast)$ are pixel coordinates of the pixel with the closest direction, $r_m = 16$ and $c_m = 32$ are margins in which we search, set experimentally to values that let always find the globally optimal pixel in the training dataset.
We also check if the distance to the point is consistent with the range measurement to account for possible occlusions stemming from the motion compensation.
The final mask is an image with pixels equal to 0 where no points were projected and 1 otherwise.

\begin{figure}[!t!]
    \centering
    \includegraphics[width=\columnwidth]{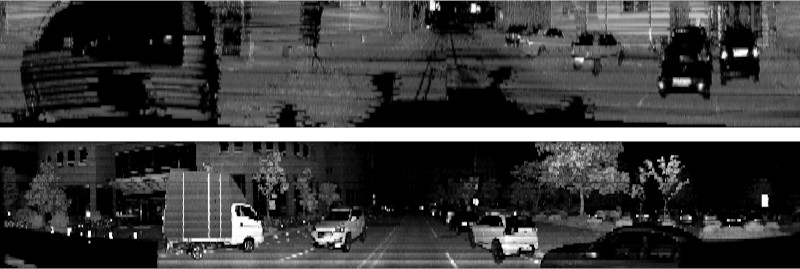}
    \caption{Visual comparison of intensity images quality from Velodyne HDL-64E (top) and Ouster OS1-64 (bottom) showing that using intensity data from the KITTI dataset might not bring significant benefits.}
    \label{fig:int_images}
\end{figure}

Laser scans from the KITTI~\cite{kitti} dataset required some additional processing, as in this case only motion compensated point clouds are available.
Thus, we perform angular bilinear interpolation to form an intensity image from measurements in directions not forming a regular grid.
For every direction on the regular grid, we select 4 nearest neighbors that fall into bins spanning from the current direction on the grid to the nearest directions on the grid.
Every neighbor is selected from a different quadrant of the image plane around the point.
Then, a horizontal angular interpolation is performed on the pair of upper and the pair of lower points separately to compute two points with the same horizontal angle as the points on the grid.
Finally, a vertical interpolation is done to compute range and intensity for the target direction.
Unfortunately, the longer laser wavelength in HDL-64E, the lack of raw measurements, and necessary
interpolations resulted in a degraded quality of the intensity images from KITTI (Fig.~\ref{fig:int_images}) that might be hindering their use to produce visual descriptors.

\subsection{DNN architecture}
\label{sec:arch}

\begin{figure}[!t!]
    \centering
    \includegraphics[width=\columnwidth]{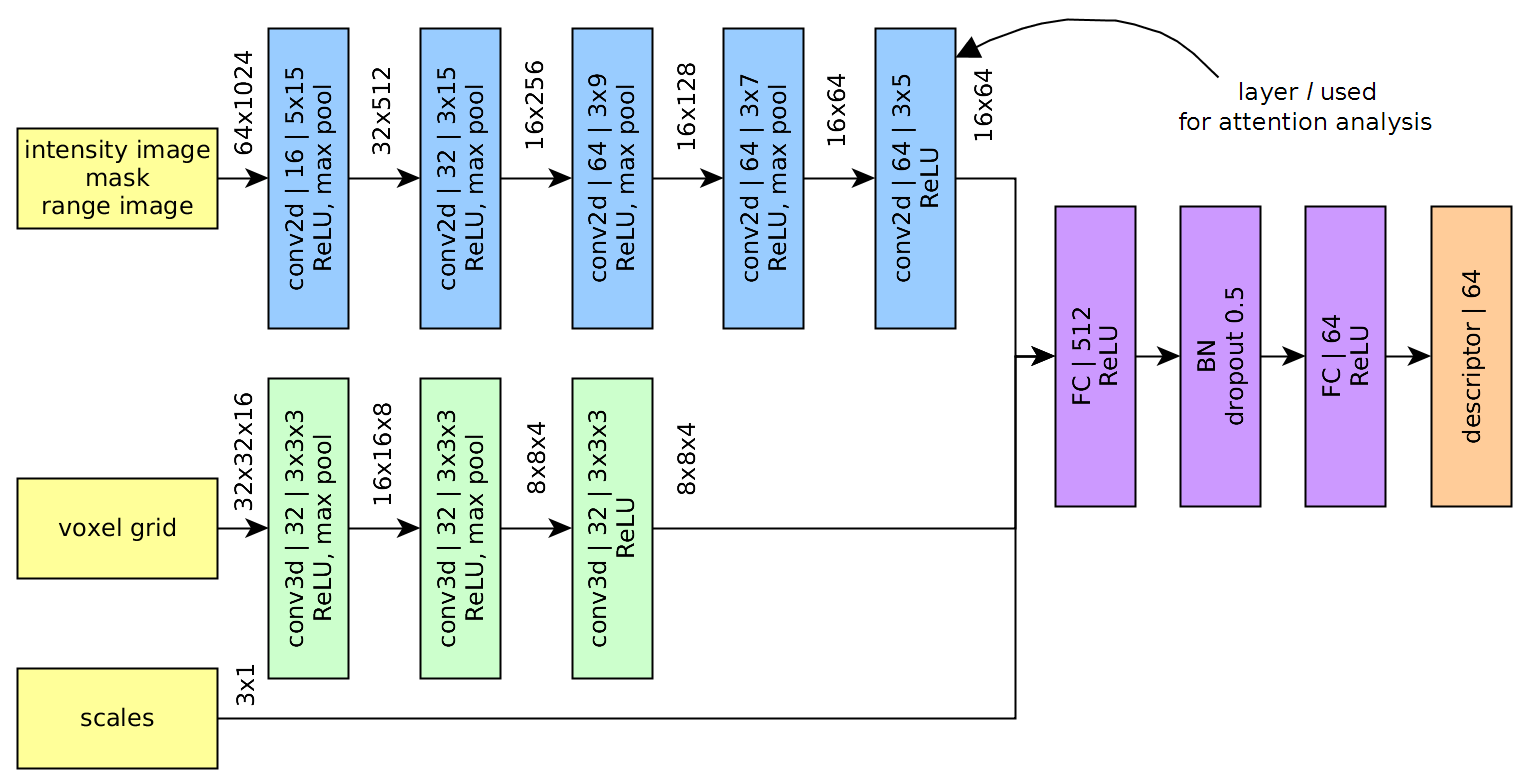}
    \caption{DNN architecture that combines visual (LiDAR intensity) and geometrical information to produce a segment descriptor. Two convolutional branches are merged using a fully connected layer.}
    \label{fig:dnn}
\end{figure}

The architecture of the DNN used to produce segment descriptors is depicted in Fig.~\ref{fig:dnn}.
It consists of two branches that are merged down-stream the computations: a geometry branch processing voxels that is the same as in SegMap, and a proposed visual branch processing intensity images.
An input to the visual branch is composed of three layers: intensity image, range image, and segment mask concatenated into a single 3 channel tensor.
The intensity image is normalized to have mean value 0 and standard deviation 1 across the whole training dataset, range image is normalized to have mean value 0 for all pixels belonging to the segment and standard deviation equal to 1, and mask values are 1 for pixels belonging to the segments and 0 otherwise.
The mask tells the DNN what it should describe and range information gives additional hints about the boundaries of objects.
By feeding the whole image instead of only a segment part, we enable the DNN to leverage the context of the segment, because its neighborhood is visible.
Typically to image processing DNNs, convolutions compress the information into higher-level features.
Finally, outputs from both branches are concatenated and the descriptor of 64$\times$1 size is computed after a fully connected layer.

\subsection{Learning}
\label{sec:learn}
The same as in~\cite{Dube2020}, we cast the description task as a classification problem, where each segment represents a different class.
Due to a large number of classes comparing to the number of training examples, the DNN produced useful descriptors without overfitting to the specific features of the classes.
We augmented training intensity images by exploiting their circular nature and randomly rotating the image around the vertical axis.
To maximize the number of distinct training examples, each segment observation was assigned an intensity image with the same timestamp during training.
Whereas during testing, we always selected the view where the segment was the most visible so far, i.e. the mask was the largest among already collected scans.
In both, training and testing, we reject images whose mask area is smaller than 50 pixels.

For DNN's parameter optimization we used Adam optimizer, batch size of 8, learning rate 0.0001, and trained for 256 epochs, selecting a model with the highest validation accuracy.

\subsection{Attention analysis}
\label{sec:attention}
There is a number of papers describing algorithms for DNN attention analysis for the classification problem~\cite{Wang2020}, but the authors were unable to find a method that is suitable when the network's output is a descriptor.
For the classification problem, ScoreCam~\cite{Wang2020} is a viable solution that does not exhibit problems with visually noisy results as the gradient-based methods~\cite{Selvaraju2017} and is relatively easy to implement.
ScoreCam computes attention heatmaps by analyzing the last layer of the DNN that has spatial dimensions (layer $l$, c.f. Fig.~\ref{fig:dnn}),
where the information is compressed the most, but the spatial structure is conserved.
For each channel in this layer, it calculates its importance by evaluating a score for the target class $c$.
The weight of the $k$-th channel is a softmax output for the target class computed by doing a forward pass with a masked input image, thus the higher the probability of the target class, the higher the weight of the channel:
\begin{equation}
    w_k^l = f(X \circ M_k^l)[c],
\end{equation}
where $[c]$ is a result for the target class $c$, $\circ$ is the element-wise product, $f(\cdot)$ denotes the forward pass, and $X$ is the input image.
The mask $M_k^l$ is calculated by upsampling activations in this channel to the size of the network's input and normalizing them:
\begin{equation}
    M_k^l = \frac{\mathrm{up}(A_k^l) - \min \left( \mathrm{up}(A_k^l) \right)}{\max \left( \mathrm{up}(A_k^l) \right) - \min \left( \mathrm{up}(A_k^l) \right)},
\end{equation}
where $A_k^l$ denotes the $k$-th activation map for the layer $l$ and $\mathrm{up}(\cdot)$ is the operator of upsampling.
This way the mask highlights image parts that were important for the activations in this channel while suppressing other parts.
The final heatmap $H^l$ is produced by multiplying channel activations with corresponding weights, upsampling to the size of the input, summing, removing negative values, and normalizing:
\begin{align}
    \tilde{H}^l = \max \left( \sum_k w_k^l \cdot \mathrm{up}(A_k^l), 0 \right), \nonumber \\
    H^l = \frac{\tilde{H}^l - \min \left(\tilde{H}^l \right)}{\max \left( \tilde{H}^l \right) - \min \left( \tilde{H}^l \right)}
\end{align}

With descriptors as output of the DNN the problem stems from the lack of a target class.
To deal with this problem, we decided to weight channels on the basis of similarity to the descriptor computed using non-masked input.
This way we measure how much the considered channel contributes to the descriptor and the smaller the difference the bigger the weight.
Denoting $\mathbf{d} = f(X)$ the descriptor for unmodified input, we compute the weights as:
\begin{equation}
    w_k^l = \frac{a}{|f(X \circ M_k^l) - \mathbf{d} |},
\end{equation}
where $a$ normalizes weights to sum up to 1.

\section{PERFORMANCE ANALYSIS}
\label{sec:perf}
We tested our solution using two datasets with different types of environments and sensors.
The first one is MulRan \cite{mulran} recorded in a lower density urban area near Daejeon in Korea with Ouster OS1-64 LiDAR, and the second one is KITTI \cite{kitti} recorded in a residential area near Karlsruhe in Germany with Velodyne HDL-64E.
In both cases, we used two sequences for training purposes, namely DCC01 and KAIST01 for MulRan, and 05 and 06 for KITTI.
We wanted to compare different types of sensors, hence our choice was MulRan that used Ouster sensor that operates in a different IR wavelength than most of the other scanners, and KITTI to enable comparison with SegMap and SegMatch.

\begin{figure}[htb!]
    \centering
    \includegraphics[width=\columnwidth]{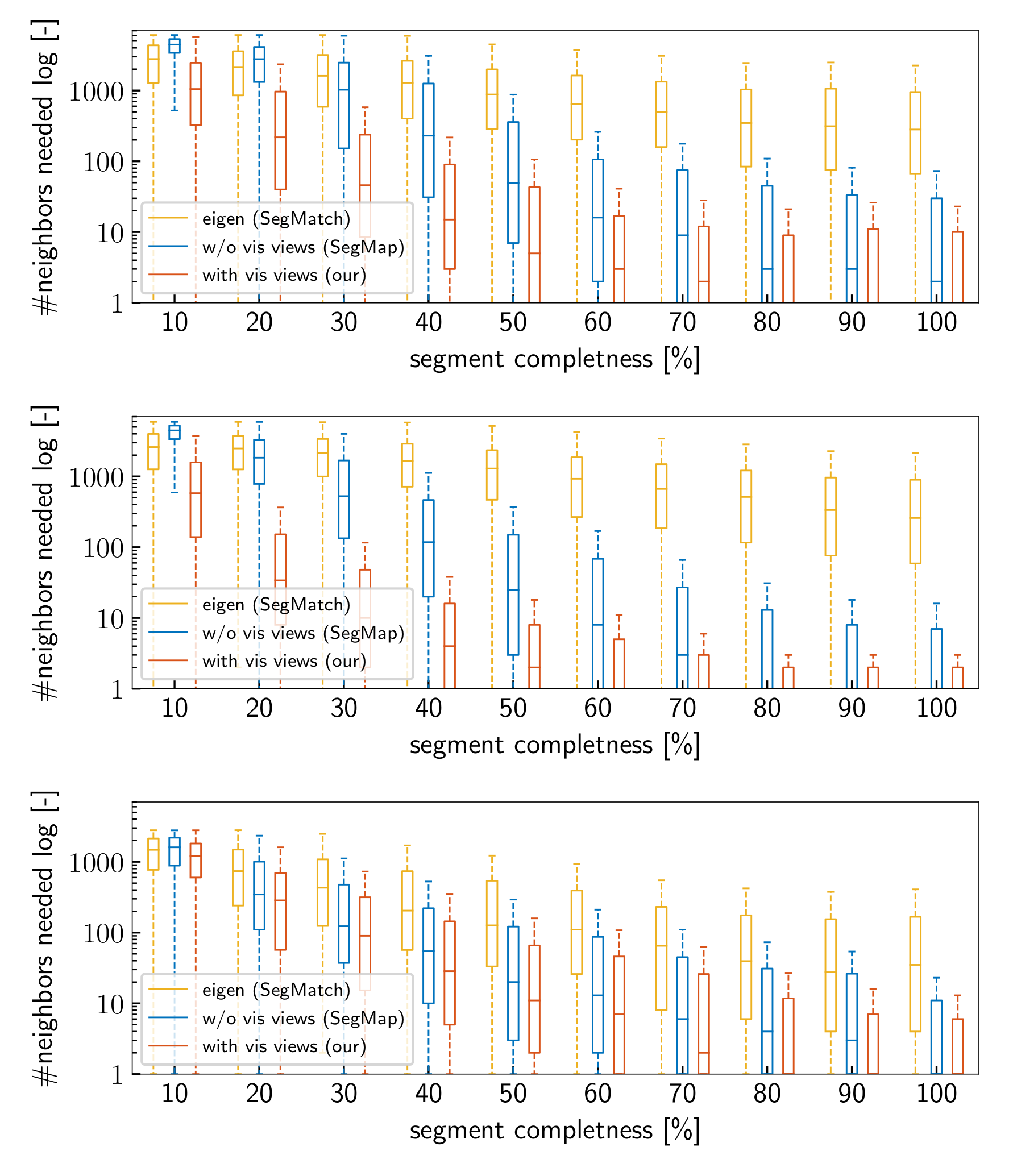}      
    \caption{Analysis of the descriptiveness of the proposed solution comparing to geometrical descriptors and eigen-based features on MulRan DCC03 (top), MulRan KAIST02 (middle), and KITTI 00 (bottom).}
    \label{fig:knn}
\end{figure}

We evaluated the performance of the proposed solution using complete sequences from both datasets that were not used during training.
For MulRan it were DCC03 and KAIST02 that have multiple loops and no large ground truth pose gaps (which appear in e.g. DCC02).
For KITTI it was the 00 sequence, that is long enough, has multiple loops, and the ground truth trajectory is accurate (as opposed to e.g. 08).
It gave us 5376 (52589 views), 5289 (54991 views), and 2659 (32073 views) testing segments, respectively, for DCC03, KAIST02, and KITTI 00.
Segments from the target map are retrieved for matching using nearest neighbors in the space of descriptors.
Thus, to evaluate the performance of our descriptors, we used the same measure as used in~\cite{Dube2020,Dube2017}, calculating how many nearest neighbors are necessary to retrieve a positive match, i.e. another observation of the same segment, excluding observations from the same sequence of observations.
We call this value segment's {\em rank} and investigate it as a function of segment's {\em completeness}, i.e. ratio of the size of a point cloud representing segment at a given time instance to the size after all observations were merged.
Segment completeness changes naturally when the vehicle moves in the environment and new measurements are incorporated into the segment representation.
Additionally, we compare our results to hand-crafted features based on the eigendecomposition of point clouds, used in SegMatch (denoted as \emph{eigen}).
Except for SegMatch and the baseline SegMap, we were unable to directly compare with other systems, either because they are not segment-based (e.g. \cite{pointnetvlad,Chen2020,Guo2019}), or there is no open source code available, as in~\cite{Tinchev2019}.
Figure~\ref{fig:knn} depicts results for different stages of segment completeness for the DCC03, KAIST02, and KITTI 00 sequence.
They indicate that descriptors with visual views outperform their purely geometric counterparts in all intervals.
It is worth noting that from 80\% of completeness, more than half of the descriptors have a positive match as the first nearest neighbor.
This fact is especially important because, during on-line operation, usually complete or almost complete segments are used.
As expected, the lowest gain is observed for the KITTI dataset containing scans from Velodyne HDL-64E which provides intensity images of considerably worse quality than those from Ouster OS1-64.

\begin{figure}[!t]
    \centering
    \includegraphics[width=\columnwidth]{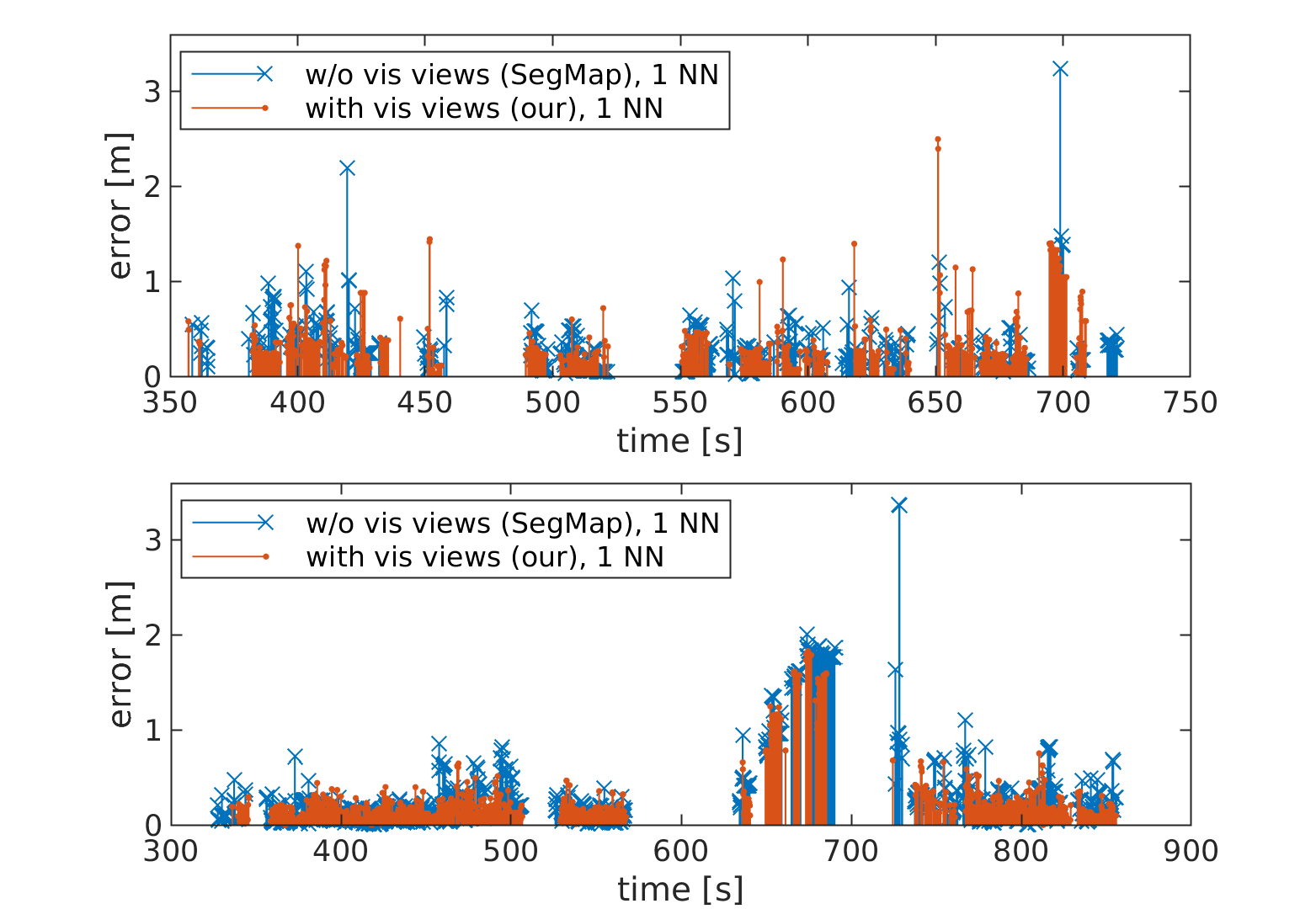}   
    \caption{Errors of relative positions (translational errors) computed for the recognized loop closures along the DCC03 (top) and KAIST02 (bottom) trajectories. Loop closures are uniformly distributed along the entire trajectories for both methods but maximum errors are smaller for our.}
    \label{fig:res_error}
\end{figure}

\begin{figure}[!t]
    \centering
    \includegraphics[width=\columnwidth]{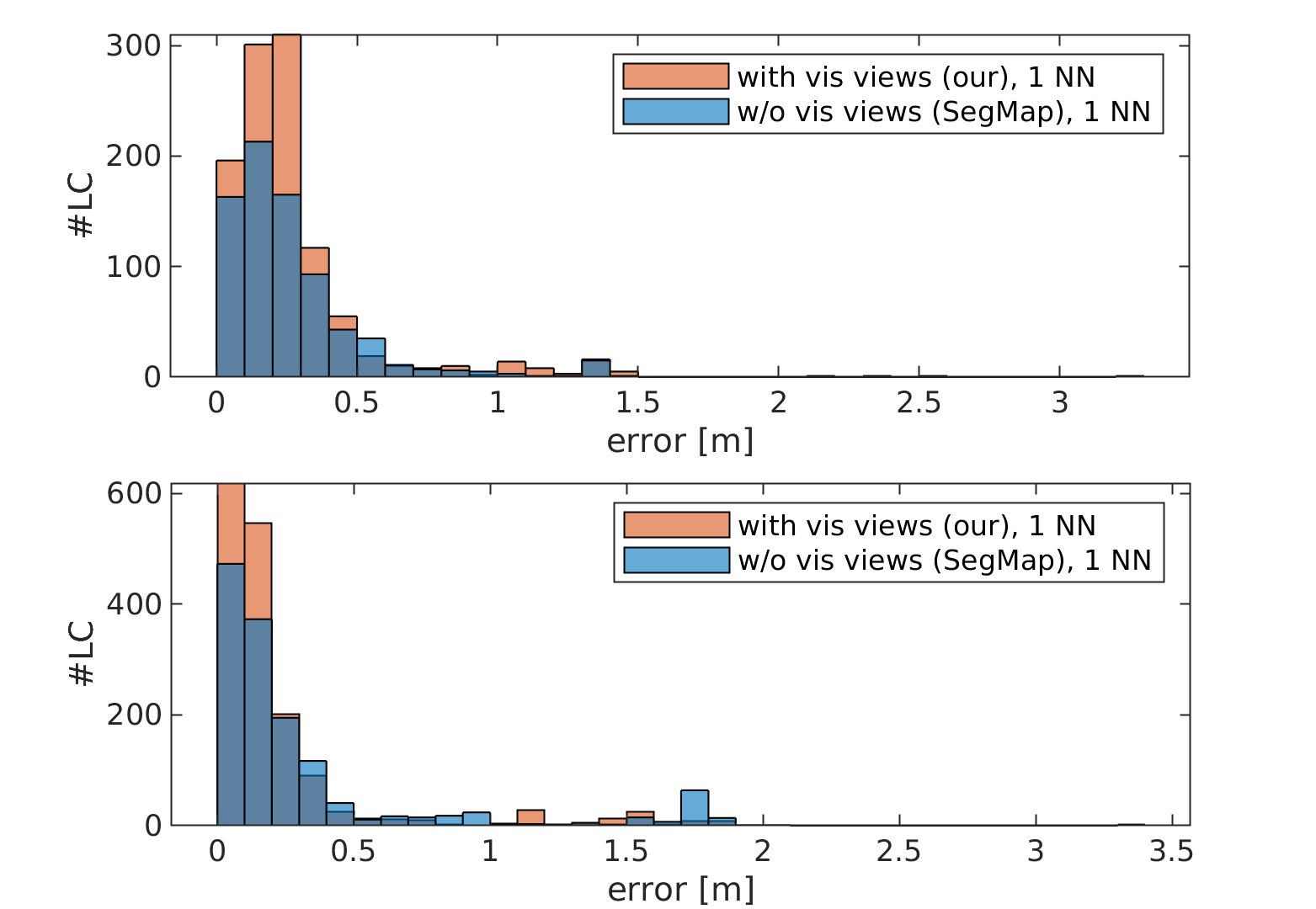}    
    \caption{Histograms of errors in the estimates position (translational errors) computed for the recognized loop closures along the DCC03 (top) and KAIST02 (bottom) trajectories. 
    Darker blue color denotes the overlapping bars of both methods. It is clearly visible that our method produces more loop closures of high position accuracy.}
    \label{fig:res_hist}
\end{figure}

\begin{table}[!htpb]
    \renewcommand{\arraystretch}{1}
    \caption{Results of loop closures. Numbers in red indicate incorrectly detected loop closures that \\[-1mm]  are unacceptable in SLAM.}
    \label{tab:res}
    \centering
    \begin{tabular}{c|l|rr|r}
        \hline
        \multicolumn{1}{c|}{seq.} & \multicolumn{1}{c|}{descriptor} & \multicolumn{1}{c}{\#corr.} & \multicolumn{1}{c|}{\#incorr.} & \multicolumn{1}{c}{error [m]} \\ \hline
        \multirow{4}{*}{\makecell{MulRan\\DCC03}} & w/o vv (SegMap), 25 NN & 1405 & \textcolor{red}{3} & 0.36 \\
                                 & with vv (our), 25 NN & 1330 & \textcolor{red}{135} & 4.97 \\
                                 & w/o vv (SegMap), 1 NN & 764 & 0 & 0.27 \\
                                 & with vv (our), 1 NN & 1076 & 0 & 0.27 \\
        \hline
        \multirow{4}{*}{\makecell{MulRan\\KAIST02}} & w/o vv (SegMap), 25 NN & 2081 & \textcolor{red}{141} & 14.69 \\
                                 & with vv (our), 25 NN & 1906 & \textcolor{red}{163} & 15.65 \\
                                 & w/o vv (SegMap), 1 NN & 1402 & 0 & 0.33 \\
                                 & with vv (our), 1 NN & 1620 & 0 & 0.23 \\
        \hline
        \multirow{4}{*}{\makecell{KITTI\\00}} & w/o vv (SegMap), 25 NN & 423 & 0 & 0.75 \\
                             & with vv (our), 25 NN & 469 & 0 & 0.75 \\
                             & w/o vv (SegMap), 1 NN & 256 & 0 & 0.69 \\
                             & with vv (our), 1 NN & 205 & 0 & 0.67 \\
        \hline
    \end{tabular}  
\end{table}

\begin{figure}[!htpb]      
    \centering
    \includegraphics[width=\columnwidth]{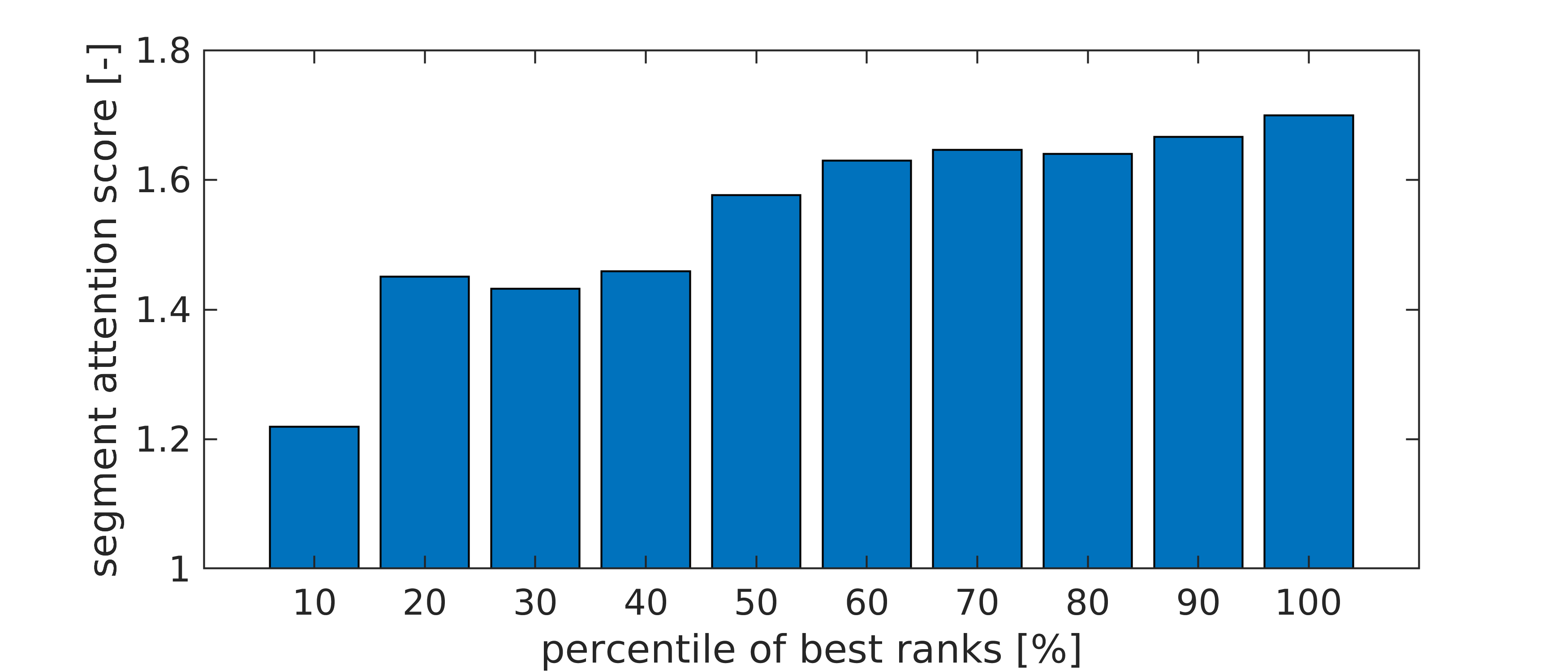}     
    \caption{Segment attention score as a function of performance of correct segment retrieval. The better the rank, the less attention is placed on the segment itself and the more on its context. }
    \label{fig:attention}
\end{figure}

To show the application potential of the new descriptors, we analyzed the number and quality of loop closures produced using them.
In most cases, including even one incorrectly recognized loop closure in SLAM optimization has a strong negative impact on the estimated trajectory \cite{sensors2021}.
Therefore, place recognition systems should avoid those, even at a cost of fewer correctly recognized places.
During initial experiments, with 25 nearest neighbors fetched for every segment in the local map (denoted \emph{25 NN}), it turned out that both solutions, without and with visual views, produced such incorrect recognitions. 
However, we observed that in most cases it is sufficient to get just the first nearest neighbor for our descriptor.
Inspired by Lowe's criterion on matching SIFT descriptors \cite{lowe2004}, we proposed to accept only match candidates which are the first nearest neighbor and for which the distance of descriptors multiplied by 1.2 is smaller than the distance to the second nearest neighbor (denoted \emph{1 NN}).
Using this criterion we eliminated incorrect recognitions while preserving a high number of correct ones.
Plots of translational errors in time are presented in Fig.~\ref{fig:res_error}, while a histogram of these errors is shown in Fig.~\ref{fig:res_hist}.
The mean position errors and numbers of recognized locations for all considered sequences are gathered in Tab.~\ref{tab:res}.
For this analysis we assumed that recognitions with a translational error greater than 5 m are incorrect.
The time plots of translational error qualitatively show that loop closures for our version are approximately uniformly distributed along the entire trajectories and are not focused in one part.
They also depict that maximum errors are smaller for our version.
The histograms show that for the MulRan sequences our method yields a higher number of loop closures that are very accurate, which is beneficial to SLAM.  
The quantitative results demonstrate also that for SLAM applications the \emph{25 NN} version is not suitable because of the incorrect recognitions.
For the \emph{1 NN} version and Ouster OS-64 (MulRan) our solution yields better results in terms of the number of correct recognitions while being better or comparable in terms of the mean position error.
The results for Velodyne HDL-64E (KITTI) are inconclusive, which we attribute to the considerably lower quality of the synthesized intensity images.

\begin{figure*}[!t!]    
    \centering
    \includegraphics[width=\textwidth]{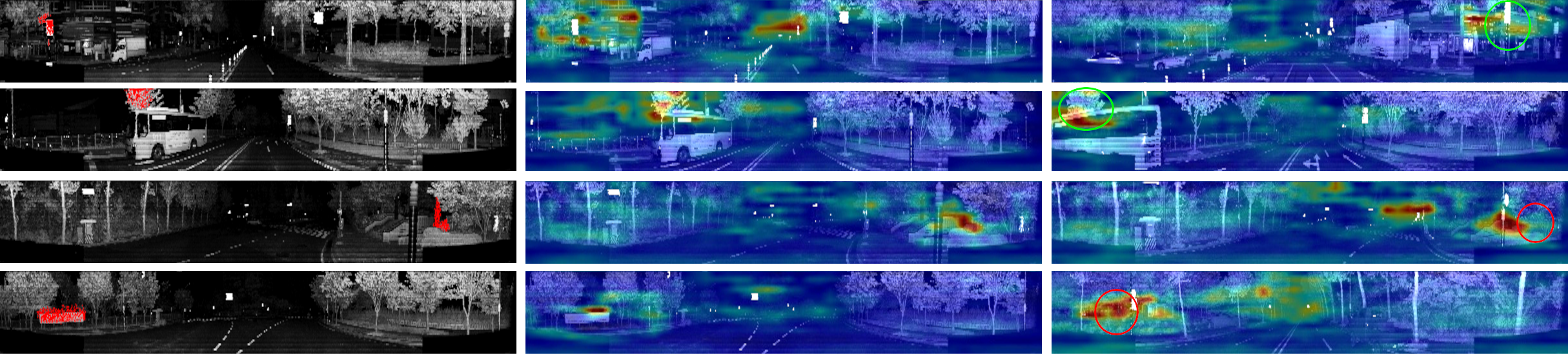}     
    \caption{Examples of first nearest neighbors retrieved from the target map as query seqment views (left), attention heatmaps for query segments (middle), and attention heatmaps for matched segments (right). Two first examples depict correctly matched segments (circled in green) from top 10\% ranks, whereas two last depict incorrectly matched ones (circled in red) with correct segments being distant in the descriptor space from bottom 10\% ranks.}
    \label{fig:attention2}
\end{figure*}

In terms of inference time, our solution is slightly faster than the original one (143 ms vs 155 ms on average using GTX 1080 Ti) thanks to a smaller number of described segments due to rejection of segments with  a too small mask.
The most time-consuming operation is the insertion of new visual views to the local map that includes finding the best visual view for every segment and takes 313 ms on average.

We use the attention analysis mechanism to demonstrate how important it is to take into consideration the visual context surrounding the segment on the image.
To show this effect quantitatively, we compute every segment's attention score as a ratio of the mean attention of pixels belonging to the mask and its nearest neighborhood in the intensity image, to the mean attention for all other pixels.
Figure~\ref{fig:attention} plots the attention score for 10 bins of segments, sorted according to their rank (the same rank as used during the performance analysis, c.f. Fig. \ref{fig:knn}).
There is a trend showing that the better the rank, the less attention is placed on the segment itself and the more on its context.
Figure~\ref{fig:attention2} visualizes attention heatmaps for two segments from the top 10\% ranks and thus being correctly associated (top rows), and two others from the bottom 10\% ranks that were mismatched (bottom rows).
There is a visible shift of attention in the DNN from the segment to the surrounding context for the correct associations, whereas in the incorrectly associated examples the DNN focused on particular objects.

\addtolength{\textheight}{-2cm}

\section{CONCLUSIONS}
We presented research aimed at improving loop closing based on the concept of geometric segments, making it possible to consider the visual context that surrounds these segments.
This context gives the segment descriptors much more descriptive power.
While there are many objects of similar shapes in real-world outdoor scenes (e.g. cars),
a combination of the segment's geometry, its texture, and other textures in the neighborhood is intuitively much more unique.
We verified this conjecture in two ways: (i) by demonstrating quantitatively on the MulRan and KITTI datasets that our new descriptors are more robust in matching than the purely geometric ones from SegMap,
(ii) by showing through DNN attention analysis that the visual context indeed matters when learning the descriptors.
Moreover, we demonstrated the processing pipeline that exploits the intensity readouts of a modern LiDAR in a way similar to passive camera images.
The new DNN architecture combines geometric and intensity data at the feature level, producing compact descriptors.
Having SLAM applications in mind and exploiting better descriptive power of our solution, we proposed a different method for selecting potential segment matches that eliminates incorrect loop closure detections that could deteriorate pose and map estimates.
The presented results are considerably better for Ouster OS1-64 due to good quality intensity images and could be further improved using the latest technology LiDAR with 128 beams \cite{ouster} and ambient images instead of the intensity images.
This possibility will be investigated in our future work.









\bibliographystyle{IEEEtran}
\bibliography{IEEEabrv,segments}

\end{document}